\documentclass[final,3p,times]{elsarticle}                    
\usepackage{graphicx}
\usepackage{enumerate}
\usepackage{url}
\usepackage{lscape}
\usepackage{subfigure, graphicx}
\begin{document}

\begin{frontmatter}
\title{Information-Theoretic Measures for Objective Evaluation of Classifications
}

\author[he,hu]{Bao-Gang Hu\corref{cor1}}
\ead{hubg@nlpr.ia.ac.cn}
\author[he]{Ran~He}
\author[yuan]{XiaoTong~Yuan}

\cortext[cor1]{Corresponding author. Address: NLPR/LIAMA, Institute of Automation, Chinese Academy of Sciences, Beijing 100190, China. Tel.: +86-10-62647318, Fax: +86-10-62647458.}

\address[he]{NLPR/LIAMA, Institute of Automation, Chinese Academy of Sciences, Beijing 100190, China}
\address[hu]{Beijing Graduate School, Chinese Academy of Sciences, Beijing 100190, China.}
\address[yuan]{Department of Electronic and Computer Engineering, National University of Singapore, Singapore.}


\begin{abstract}
This work presents a systematic study of objective evaluations
of abstaining classifications using Information-Theoretic Measures
(\textbf{ITMs}). First, we define objective measures for which they
do not depend on any free parameter. This definition provides
technical simplicity for examining
``\emph{objectivity}"  or ``\emph{subjectivity}" directly to
classification evaluations. Second, we propose twenty four normalized
ITMs, derived from either mutual information, divergence, or
cross-entropy, for investigation. Contrary to conventional
performance measures that apply empirical formulas based on users'
intuitions or preferences, the ITMs are theoretically more sound for
realizing objective evaluations of classifications. We apply them to
distinguish  ``\emph{error types}" and ``\emph{reject types}" in
binary classifications without the need for input data of cost
terms. Third, to better understand and select the ITMs, we suggest
three desirable features for classification assessment measures,
which appear more crucial and appealing from the viewpoint of
classification applications. Using these features as
``\emph{meta-measures}", we can reveal the advantages and
limitations of ITMs from a higher level of evaluation knowledge.
Numerical examples are given to corroborate our claims and compare
the differences among the proposed measures. The best measure
is selected in terms of the meta-measures, and its specific
properties regarding error types and reject types are
analytically derived.
\end{abstract}

\begin{keyword}
Abstaining classifications \sep error types \sep reject types \sep
entropy \sep similarity \sep objectivity
\end{keyword}
\end{frontmatter}

\section{Introduction}
The selection of evaluation measures for classifications has
received increasing attentions from
researchers on various application fields \cite{Ling:2003}
\cite{Beg:2005}\cite{Japkowicz:2006}\cite{Pietraszek:2007}\cite{Lavesson:2007}
\cite{Vanderlooy:2008}\cite{Hand:2009}.
It is well known that evaluation measures, or criteria, have a
substantial impact on the quality of classification performance. The
problem of how to select evaluation measures for the overall quality
of classifications is difficult, and there appears no universal
answer to this.

Up to now, various types of evaluation measures have been used in
classification applications. Taking a binary classification as an
example, more than thirty metrics have been applied for assessing
the quality of classifications and their algorithms as given in
Table 1 of Lavesson and Davidsson's paper \cite{Lavesson:2007}. Most
of the metrics listed in this table can be considered a type of
performance-based measures. In practice, other types of evaluation
measures, such as Information-Theoretic Measures (\textbf{ITMs}),
have also commonly been used in machine learning
\cite{Yao:1999}\cite{Principe:2000}. The typical information-based
measure used in classifications is the cross entropy
\cite{Bishop:1995}. In a recent work \cite{Hu:2008}, Hu and Wang
derived an analytical formula of the Shannon-based mutual
information measure with respect to a confusion matrix. Significant
benefits were derived from the measure, such as its generality even
for cases of classifications with a reject option, and its
objectivity in naturally balancing performance-based measures that
may conflict with one another (such as precision and recall). The
objectivity was achieved from the perspective that an
information-based measure does not require knowledge of cost terms
in evaluating classifications. This advantage is particularly
important in studies of abstaining classifications
\cite{Temanni:2007}\cite{Pietraszek:2007}
and cost sensitive learning
\cite{Domingos:1999}\cite{Elkan:2001}, where
cost terms may be required as input data for evaluations. Generally,
if no cost terms are assigned to evaluations, it implies that the
zero-one cost functions are applied \cite{Duda:2001}. In such
situations, classification evaluations without a reject option may
still be applicable and useful in class-balanced datasets.
Problematic, or unreasonable, results will be obtained for
evaluations in situations where classes are highly skewed in the
datasets \cite{Japkowicz:2006} if no specific cost terms are given.

In this work, for simplifying discussions, we distinguish, or decouple,
two study goals in evaluation studies, namely, evaluation of classifiers
and evaluation of classifications. The former goal concerns more
about evaluation of algorithms in which classifiers applied.
From this evaluation, designers or users can select the best classifier.
The latter goal is to evaluate classification results without
concerning which classifier is applied. This evaluation aims more on
result comparisons or measure comparisons. One typical example
was demonstrated by Mackay \cite{Mackay:2003} for highlighting the difficulty
in classification evaluations.
He showed two specific confusion matrices, $\textbf{C}_D$
and $\textbf{C}_E$, in binary
classifications with a reject option:
\begin{equation}
\textbf{C}_D = \left[ {\begin{array}{*{20}c}
   {74} & {6 }  & {10 }  \\
   {0} & {9 } & {1}  \\
\end{array}} \right], \quad
\textbf{C}_E = \left[ {\begin{array}{*{20}c}
   {78} & {6 }  & {6 }  \\
   {0} & {5 } & {5}  \\
\end{array}} \right],\quad with \quad
\textbf{C} = \left[ {\begin{array}{*{20}c}
   {TN} & {FP }  & {RN }  \\
   {FN} & {TP } & {RP}  \\
\end{array}} \right],
\end{equation}
where the confusion
matrix is defined as $\textbf{C}$ in eq. (1) ,
and ``\emph{TN}", ``\emph{TP}", ``\emph{FN}", ``\emph{FP}",
``\emph{RN}", ``\emph{RP}"
represent ``\emph{true negative}" , ``\emph{true positive}",
``\emph{false negative}", ``\emph{false positive}", ``\emph{reject negative}",
``\emph{reject positive}", respectively. For the
given data, users may ask ``\emph{which measures will be
proper for ranking them}". If directly applying
``\emph{True Positive Rate-False Positive Rate}"  curve (also called ROC)
or ``\emph{Precision-Recall}" curve, one may conclude that
the performance of $\textbf{C}_E$  is better than that of $\textbf{C}_D$ .
This conclusion is proper since the two sets of data
share the same reject rate (=11\%).
Generally, ``\emph{Error-Reject}" curve is mostly adopted in
abstaining classifications. Based on this evaluation
approach, one may consider the performances of two
classifications have no difference because they show
the same error rate (=6\%)
and reject rate. Mackay \cite{Mackay:2003} first suggested applying
mutual-information based measure in ranking classifications, and through which
Hu and Wang (referring to M5-M6 in Table 3,
\cite{Hu:2008}) observed that $\textbf{C}_D$  is better than
$\textbf{C}_E$. If reviewing the two matrices carefully with respect to
imbalanced classes,
one may agree with the observation because the small
class in $\textbf{C}_D$  receives more correct classifications than
that in $\textbf{C}_E$.

We consider the example designed by Mackay \cite{Mackay:2003}
is quite stimulating for study of abstaining classification evaluations.
The implications of the example form the motivations of the present work
on addressing three related open problems,
which are generally overlooked in the study of classification
evaluations as follows:
\begin{itemize}
\item[I.] How to define ``\emph{proper}" measures in terms of high-level knowledge for 
abstaining classification evaluations?%
\item[II.] How to conduct an objective evaluation of classifications without using cost terms?%
\item[III.] How to distinct or rank ``\emph{error types}" and ``\emph{reject types}" in classification evaluations?%
\end{itemize}

Conventional binary classifications usually distinguish two types of
misclassification errors \cite{Duda:2001}\cite{Mackay:2003} if they
result in different losses in applications. For example, in medical
applications, ``\emph{Type I Error}" (or ``\emph{false positive}")
can be an error of misclassifying a healthy person to be abnormal,
such as cancer. On the contrary, ``\emph{Type II Error}"(or
``\emph{false negative}") is an error where cancer is not detected
in a patient. Therefore, ``\emph{Type II Error}" is more
costly than ``\emph{Type I Error}". Based on the same
reason for identifying ``\emph{error types}" in binary
classifications, there is a need for considering ``\emph{reject
types}" if a reject option is applied. Of the existing measures, we
consider information-theoretic measures to be most promising in
providing ``\emph{objectivity}" in classification evaluations. A
detailed discussion on the definition of ``\emph{objectivity}" is
given in Section 3. This work is an extension of our previous study
\cite{Hu:2008}. However, the work aims at a systematic investigation
of information measures with specific focus on ``\emph{error types}"
and ``\emph{reject types}". The main contribution of the work is
derived from the following three aspects:
\begin{itemize}
\item[I.] We define the ``\emph{proper}" features, also called ``\emph{meta-measures}"
, for selecting candidate measures in the context of abstaining classification
evaluations. These features will assist users in understanding
advantages and limitations of evaluation measures from a higher
level of knowledge.
\item[II.] We examine most of the existing information measures
in a systematic investigation of ``\emph{error types}" and
``\emph{reject types}" for objective evaluations. We hope that the
more than twenty measures investigated are able to enrich the
current bank of classification evaluation measures. For the
best measure in terms of the meta-measures, we present a
theoretical confirmation of its desirable properties
regarding error types and reject types.
\item[III.] We reveal the intrinsic shortcomings of information
measures in evaluations. The discussions are intended to be
applicable to a wider range of classification problems, such as
similarity ranking. In addition, we are able to employ the measures
reasonably in interpreting classification results.
\end{itemize}

To address classification evaluations with a reject option, we
assume that the only basic data available for classification
evaluations is a confusion matrix, without input data of cost terms.
The rest of this letter is organized as follows. In Section 2, we
present related work for the selection of evaluation measures. For
seeking ``\emph{proper}" measures, we propose several desirable
features in the context of classifications in Section 3. Three
groups of normalized information measures are proposed along with
their intrinsic shortcomings in Sections 4 to 6, respectively.
Several numerical examples, together with discussions, are given in
Section 7. Finally, in Section 8 we conclude the work.

\section{Related Work}
In classification evaluations, a measure based on classification
accuracy has traditionally been used with some success in numerous
cases \cite{Duda:2001}. This measure, however, may suffer serious
problems in reaching intuitively reasonable results from certain
special cases of real-world classification problems
\cite{Japkowicz:2006}. The main reason for this is that a single
measure of accuracy does not take into account error types.

To overcome the problems of accuracy measures, researchers have
developed many sophisticated approaches for classification
assessment\cite{Ferri:2009}\cite{Japkowicz:2011}. Among these, two commonly-used approaches are ROC
(Receiver Operating Characteristic) curves and AUC (Area under
Curve) measures
\cite{Ling:2003}\cite{Fawcett:2006}. ROC curves
provide users with a very fast evaluation approach via visual
inspections, but this is only applicable in limited cases with
specific curve forms (for example, when one curve is completely
above the other). AUC measures are more generic for ranking
classifications without constraints on curve forms. In a study of binary
classifications, a formal proof was given by Ling et al.
\cite{Ling:2003} showing that AUC is a better measure than accuracy
from the definitions of both statistical consistency and
discriminancy.
Sophisticated AUC measures were reported recently
for improving robustness \cite{Vanderlooy:2008}
and coherency \cite{Hand:2009} of classifiers.
Drummond and Holte \cite{Drummond:2006} proposed
a visualization technique called
``\emph{Cost Curve}", which is able to take into account of
cost terms for showing confidence intervals
on classifier's performance.
Japkowicz \cite{Japkowicz:2006} presented convincing
examples showing the shortcomings of the existing evaluation
methods, including accuracy, precision vs. recall, and ROC
techniques. The findings from the examples further confirmed the
need for methods using measure-based functions
\cite{Andersson:1999}. The main idea behind measure-based functions
is to form a single function with respect to a weighted summation of
multiple measures. The measure function is able to balance a
trade-off among the conflicting measures, such as precision and
recall. However, the main difficulty arises in the selection of
balancing weights for the measures \cite{Lavesson:2007}. In most
cases, users rely on their preferences and experiences in assigning
the weights, which imposes a strong degree of subjectivity on the
evaluation results.

Classification evaluations become more complicated if a classifier
abstains from making a prediction when the outcome is considered
unreliable for a specific sample. In this case, an extra class,
known as the ``\emph{reject}" or ``\emph{unknown}" class, is added
to the classification. In recent years, the study of abstaining
classifiers has received much attention
\cite{Stefano:2000}\cite{Landgrebe:2006}\cite{Temanni:2007}\cite{Pietraszek:2007}\cite{Iannello:2009}.
With complete data of a full cost matrix, they were able to assess
the classifications. If one term of the cost matrix was missing,
such as on a reject cost term, the approaches for classification
evaluations generally failed. Moreover, because in most situations
the cost terms are given by users, this approach is basically a
subjective evaluation in applications. Vanderlooy et al.
\cite{Vanderlooy:2009} further investigated the ROC isometrics
approach which does not rely on information from a cost matrix. This
approach, however, is only applicable to binary classification
problems.

A promising study of objective evaluations of classifications is
attributed to the introduction of information theory. Kvalseth
\cite{Kvalseth:1987} and Wickens \cite{Wickens:1989} derived
normalized mutual information (\textbf{NMI}) measures in relation to
a contingency table. Further pioneering studies on the
classification problems were conducted by Finn \cite{Finn:1993} and
Forbes \cite{Forbes:1995}. Forbes \cite{Forbes:1995} discussed the
problem that NMI does not share a monotonic property with the other
performance measures, such as accuracy or F-measure.
Several different definitions for information measures have
been reported in studies of classification assessment, such as
information scores by Kononenko and Bratko \cite{Kononenko:1991} and
KL divergence by Nishii and Tanaka \cite{Nishii:1999}. Yao, et al.
\cite{Yao:1999} and Tan, et al. \cite{Tan:2004} summarized many
useful information measures for studies of associations and
attribute importance. Significant efforts were made on discussing
the desired properties of evaluation measures \cite{Tan:2004}.
Principe, et al. \cite{Principe:2000} proposed
a framework of information theoretic learning (\textbf{ITL}) that
included supervised learning as in classifications. Within this
framework, the learning criteria were the mutual information defined
from the Shannon and Renyi entropies. Two quadratic divergences,
namely, the Euclidean and Cauchy-Schwartz distances were also
included.

From the perspective of information theory,
Wang and Hu \cite{Wang:2009} derived for the first time the
nonlinear relations between mutual information and the conventional
performance measures (accuracy, recall and precision) for
binary classification problems. They \cite{Hu:2008} extended the investigation into
abstaining classification evaluations for multiple classes. Their
method was based solely on the confusion matrix.
For gaining the theoretical properties, they
derived the extremum theorems concerning mutual
information measures. One of the important findings from the local
minimum theorem is the theoretic revelation of the non-monotonic
property of mutual information measures with respect to the diagonal
terms of a confusion matrix. This property may cause irrational
evaluation results from some data in classifications. They confirmed
this problem by examining specific numerical examples.
Theoretical investigations are still missed for other information measures,
such as divergence-based and cross-entropy based ones.

\section{Objective Evaluations and Meta-Measures}
This work focuses on objective evaluations of classifications. While
Berger \cite{Berger:2006} stressed four points from a philosophical
position for supporting objective Bayesian analysis, it seems that
few studies in the literature address the ``\emph{objectivity}"
issue in the study of classification evaluations.
Some researchers \cite{Tan:2004} may call their measures to be
objective ones without defining them formally.
Considering that
``\emph{objectivity}" is a more philosophical concept without a well
accepted definition, we propose a scheme for defining
``\emph{objective evaluations}" from the viewpoint of practical
implementation and examination.

\textbf{Definition 1. Objective evaluations and measures}. An objective
evaluation is an assessment expressed by a function
that does not contain any free parameter. This function is
called an objective measure.

\textbf{Remark 1}. When a free parameter is used to
define a measure, it usually carries a certain degree of subjectivity
in evaluations. Therefore, according to this definition,
a measure based on cost terms \cite{Duda:2001} as free parameters
does not lead to an objective evaluation. Definition 1 may be
conservative, but nevertheless,
provides technical simplicity for examining
``\emph{objectivity}" or ``\emph{subjectivity}" directly with
respect to the existence of free parameters. In some situations, 
Definition 1 can be relaxed by including free parameters, but they all
have to be determined solely from the given dataset.

\textbf{Definition 2. Datasets in classification evaluations with a
reject option}. A reject option is sometimes considered for
classifications in which one may assign samples to a reject or
unknown class. Evaluations of classification with a reject option
apply two datasets, namely, the output (or prediction) dataset $ \{
y_k \} _{k = 1}^n $, which is a realization of discrete random
variable $Y$ valued on set $ \{ 1,2, \dots ,m+1\}$; and the target
dataset $ \{ t_k \} _{k = 1}^n \in T $ valued on set $ \{ 1,2, \dots
,m\} $; where $n$ is the total number of samples, and $m$ is the
total number of classes. A sample identified as a reject class is
represented by $y_k= m+1$.

\textbf{Remark 2}. The term ``\emph{abstaining classifiers}" has
been widely used in classification problems with a reject option
\cite{Temanni:2007}\cite{Pietraszek:2007}.
However, most studies of abstaining classifications required cost
matrices for their evaluations. The definition given above exhibits
more generic scenarios in classification evaluations, because it
does not require knowledge of cost terms for error types and reject
types.

\textbf{Definition 3. Augmented confusion matrix and its constraints
\cite{Hu:2008}}. An augmented confusion matrix includes one column
for the reject class, which is added to a conventional confusion
matrix:
\begin{equation}
\textbf{C} = \left[ {\begin{array}{*{20}c}
   {c_{11} } & {c_{12} } & {\dots} & {c_{1m} } & {c_{1(m + 1)} }  \\
   {c_{21} } & {c_{22} } & {\dots} & {c_{2m} } & {c_{2(m + 1)} }  \\
   {} & {} & {\dots} & {} & {}  \\
   {c_{m1} } & {c_{m2} } & {\dots} & {c_{mm} } & {c_{m(m + 1)} }  \\
\end{array}} \right],
\end{equation}
where $c_{ij}$ represents the sample number of the $i$th class that
is classified as the $j$th class. The row data corresponds to the
actual classes, while the column data corresponds to the predicted
classes. The last column represents the reject class. The relations and
constraints of an augmented confusion matrix are:
\begin{equation}
C_j  = \sum\limits_{j = 1}^{m + 1} {c_{ij} } ,\quad C_i>0,\quad
c_{ij} \ge 0, \quad i=1,2,\dots ,m,
\end{equation}
where $C_i$ is the total number for the $i$th class, which is
generally known in classification problems.

\textbf{Definition 4. Error types and reject types}. Following the
conventions in binary classifications \cite{Montgomery:2006},
we denote $c_{12}$ and $c_{21}$ by ``\emph{Type I Error}" and
``\emph{Type II Error}" respectively; $c_{13}$ and $c_{23}$ by
``\emph{Type I Reject}" and ``\emph{Type II Reject}" respectively.

\textbf{Definition 5. Normalized information measure}. A normalized
information measure, denoted as $NI(T,Y) \in [0,1]$, is a function
based on information theory, which represents the degree of
similarity between two random variables $T$ and $Y$.

In principle, we hope that all NI measures satisfy the three
important properties, or axioms, of metrics
\cite{Duda:2001}\cite{Li:2004}, supposing $Z$ is another random
variable:
\begin{itemize}
\item[] P1: $NI(T,Y)=1$ \emph{iff} $ T=Y$ (the identity axiom)
\item[] P2: $NI(T,Y)+NI(Y,Z) \ge NI(T,Z)$ (the triangle inequality)
\item[] P3: $NI(T,Y)=NI(Y,T)$ (the symmetry axiom)
\end{itemize}

\textbf{Remark 3}. Violations of properties of metrics are possible
in reaching reasonable evaluations of classifications. For example,
the triangle inequality and symmetry properties can be relaxed
without changing the ranking orders among classifications if their
evaluation measures are applied consistently. However, the identity
property is indicated only for the relation $T=Y$ (assuming $T$ is
padded with zeros to make it the same size as $Y$), and does not
guarantee an exact solution ($t_k = y_k$) in classifications (see
Theorems 1 and 4 given later). If a violation of metric properties
occurs, the NIs are referred to as measures, rather than metrics.

For classification evaluations, we consider the
generic properties of metrics not to be as crucial in comparisons as
certain specific features. In this work, we focus on specific
features that, though not mathematically fundamental, are more
necessary in classification applications. To select ``\emph{better}"
measures for objective evaluations of classifications, we propose
the following three desirable features together with their heuristic
reasons.

\textbf{Feature 1. Monotonicity with respect to the diagonal terms
of the confusion matrix}. The diagonal terms of the confusion matrix
represent the exact classification numbers for all the samples. Or,
they reflect the coincident numbers between $t$ and $y$ from a
similarity viewpoint. When one of these terms changes, the
evaluation measure should change in a monotonous way. Otherwise, any
non-monotonic measure may fail to provide a rational result for
ranking classifications correctly. This feature is originally proposed for
describing the strength of agreement (or similarity) if the
matrix is a contingency table \cite{Tan:2004}.

\textbf{Feature 2. Variation with reject rate}. To improve
classification performance, a reject option is often used in
engineering applications
\cite{Temanni:2007}. Therefore,
we suggest that a measure should be a scalar function on both
classification accuracy and reject rates. Such a measure could be
evaluated based solely on a given confusion matrix from a
single operating point in the classification. This is different
to the AUC measures that are based on an ``\emph{Error-Reject}"
curve \cite{Mackay:2003}\cite{Iannello:2009} from multiple operating
points.

\textbf{Feature 3. Intuitively consistent costs among error types
and reject types}. This feature is derived from the principle of our conventional
intuitions when dealing with error types in classifications. It is also
extended to reject types. Two specific intuitions are adopted
for binary classifications. First, a misclassification
or rejection from a small class will cause a greater cost than
that from a large class. This intuition represents a property
called ``\emph{within error types and reject types}". Second, 
a misclassification will produce a greater
cost than a rejection from the same class, which is
called ``\emph{between error and reject types}" property.
If a measure is able to satisfy the intuitions,
we refer to its associated costs as being ``\emph{intuitively
consistent}". Exceptions may exist to the intuitions above, but we
consider them as a very special case. 

At this stage, it is worth discussing on ``\emph{objectivity}"
in evaluations because one may doubt correctness of the intentions
above and the terms ``\emph{desirable}" or ``\emph{intuitions}" in a
study of objective evaluations. The three features seem to be
``\emph{problematic}" in terms of providing a general concept of
``\emph{objectivity}", because no human bias should be applied in
the objective judgment of evaluation results. The following discussions
justify the proposal of requiring desirable,
or proper, features for objective measures. On one hand, we
recognize that any evaluation will imply a certain degree
of ``\emph{subjectivity}", since evaluations exist only as
a result of human judgment. For examples, every selection of evaluation
measures, even of objective ones, will rely on possible sources
of ``\emph{subjectivity}" from users. On the other hand, engineering
applications do concern about objective evaluations
\cite{Forbes:1995}\cite{Tan:2004}. However, to the authors' best knowledge,
a technical definition, or criterion, seems missing for determining
objective or subjective measures in evaluations of classifications.
For overcoming possible confusion and vagueness, we set Definition 1
as a practical criterion for examining whether a classification evaluation
holds ``\emph{objectivity}" or does not. If a measure satisfies this
definition, it will always retain the property of
``\emph{objective consistency}" in evaluating the given classification
results. The three ``\emph{desirable}" features, though based on
``\emph{intuitions}" with ``\emph{subjectivity}", do not destroy
the criterion of ``\emph{objectivity}" in classification evaluations.
Therefore, it is logically correct to discuss ``\emph{desirable}" features of
objective measures as long as the measures satisfy Definition 1 for
keeping the defined ``\emph{objectivity}".

Note that all desirable features
above are derived from our intuitions on general cases of
classification evaluations. Other items may be derived for a wider
examination of features. For example, Forbes \cite{Forbes:1995}
proposed six ``\emph{constraints on proper comparative measures}",
namely, ``\emph{statistically principled, readily interpretable,
generalizable to k-class situations, not different to the special
status, reflective of agreement, and insensitive to the
segmentation}". However, we consider the three features proposed in
this work to be more crucial, especially as Feature 3 has never been
concerned in previous studies of classification evaluations.
Although Features 2 and 3 may share a similar meaning, they are
presented individually to highlight their specific concerns. We can also
call the desirable features ``\emph{meta-measures}", since these are
defined to be qualitative and high-level measures about measures. In
this work, we apply meta-measures in our investigation of
information measures. The examination with respect to the
meta-measures enables clarification of the causes of performance
differences among the examined measures in classification
evaluations. It will be helpful for users to understand advantages
and limitations of different measures, either objective- or
subjective-ones, from a higher level of evaluation knowledge.

\section{Normalized Information Measures based on Mutual
Information} All NI measures applied in this work are divided into
one of three groups, namely, mutual-information based, divergence based,
and cross-entropy based groups. In this section, we focus on the
first group. Each measure in this group is derived directly from
mutual information representing the degree of similarity between two
random variables. For the purpose of objective evaluations, as
suggested by Definition 1 in the previous section, we eliminate all
candidate measures defined from the Renyi or Jensen
entropies \cite{Tsallis:1988}\cite{Principe:2000} since they involve
a free parameter. Therefore, without adding free parameters, we only
apply the Shannon entropy to information measures \cite{Cover:2006}:
\begin{equation}
 H(Y) =  - \sum\limits_y {p(y)\log _2 p(y)},
\end{equation}
where $Y$ is a discrete random variable with probability mass
function $p(y)$. Then mutual information is defined as
\cite{Cover:2006}:
\begin{equation}
I(T,Y) = \sum\limits_t {\sum\limits_y {p(t,y)\log _2
\frac{{p(t,y)}}{{p(t)p(y)}}} },
\end{equation}
where $p(t, y)$ is the joint distribution for the two discrete
random variables $T$ and $Y$, and $p(t)$ and $p(y)$ are called
marginal distributions that can be derived from:
\begin{equation}
p(t) = \sum\limits_y {p(t,y)},p(y) = \sum\limits_t {p(t,y)}.
\end{equation}

Sometimes, the simplified notations for
$p_{ij}=p(t,y)=p(t=t_i,y=y_j)$ are used in this work. Table 1 lists
the possible normalized information measures within the
mutual-information based group. Basically, they all make use of Eq.
(5) in their calculations. The main differences are due to the
normalization schemes. In applying the formulas for calculating
$NI_k$, one generally does not have an exact $p(t, y)$. For this
reason, we adopt an empirical joint distribution defined below for
the calculations.

\textbf{Definition 6. Empirical joint distribution and empirical
marginal distributions \cite{Hu:2008}}. An empirical joint
distribution is defined from the frequency means for the given
confusion matrix, $\textbf{C}$, as:

$$
P_e (t,y) = (P_{ij})_e = \frac{1}{n}c_{ij},\quad i =
1,2,\dots,m,\quad j = 1,2,\dots,m + 1, \eqno{(7a)}
$$
where $n = \sum {C_i }$, denotes the total number of samples in the
classifications. The subscript "$e$" is given for denoting empirical
terms. The empirical marginal distributions are:

$$
P_e (t=t_i) =  {\frac{{C_i }}{n}}  ,\quad i = 1,2,\dots,m.
\eqno{(7b)}
$$
$$
P_e (y=y_j) =  {\frac{1}{n}\sum\limits_{i = 1}^m {c_{ij} } }
 , \quad j = 1,2,\dots,m + 1. \eqno{(7c)}
$$
\textbf{Definition 7. Empirical mutual information \cite{Hu:2008}}.
The empirical mutual information is given by:

$$
I_e (T,Y) = \sum\limits_t {\sum\limits_y {P_e (t,y)\log _2
\frac{{P_e (t,y)}}{{P_e (t)P_e (y)}}} }  =  \sum\limits_{i = 1}^m
{\sum\limits_{j = 1}^{m + 1} {\frac{{c_{ij} }}{n}\log _2
(\frac{{c_{ij} }}{{C_i \sum\limits_{i = 1}^m {\frac{{c_{ij} }}{n}}
}})} }. \eqno{(8)}
$$

Definitions 6 and 7 provide users with a direct means for applying
information measures through the given data of the confusion matrix.
For the sake of simplicity of analysis and discussion, we adopt the
empirical distributions, or $p_{ij}\approx P_{ij}$, for calculating
all NIs and deriving the theorems, but removing their associated
subscript "$e$". Note that the notation of $NI_2$ in Table 1 differs
from the others for calculating mutual information, where $I_M(T,Y)$
is defined as ``\emph{modified mutual information}", The calculation
of $I_M(T,Y)$ is carried out based on the intersection of T and Y.
Hence, when using Eq. (8), the intersection requires that $I_M(T,Y)$
incorporate a summation of $j$ over $1$ to $m$, instead of $m+1$.
This definition is beyond mathematical rigor, but $NI_2$ has the
same properties of metrics as $NI_1$. It was originally proposed to
overcome the problem of unchanging values in NIs if rejections are
made within only one class (referring to M9-M10 in Table 3,
\cite{Hu:2008}). The following three theorems are derived for all
NIs in this group.

\textbf{Theorem 1}. Within all NI measures in Table 1, when
$NI(T,Y)=1$, the classification without a reject class may
correspond to the case of either an exact classification
($y_k=t_k$), or a specific misclassification ($y_k \ne t_k$). The
specific misclassification can be fully removed by simply exchanging
labels in the confusion matrix.

\textbf{Proof}. If $NI(T,Y)=1$, we can obtain the following
conditions from Eq. (8) for classifications without a reject class:

$$
p_{ij}  = p(t=t_i) \approx P_e (t=t_i) = \frac{{C_i }}{n} \quad and
\quad p_{kj}=0, \quad i,j,k=1, 2, \dots ,m, \quad k \ne i.
 \eqno{(9)}
$$
These conditions describe the specific confusion matrix where only
one non-zero term appears in each column (with the exception of the
last $(m+1)$th column). When $j=i$, \textbf{C} is a diagonal matrix
for representing an exact classification ($y_k=t_k$). Otherwise, a
specific misclassification exists for which a diagonal matrix can be
obtained by exchanging labels in the confusion matrix (referring to
M11 in Table 4, \cite{Hu:2008}). $\diamondsuit$

\textbf{Remark 4}. Theorem 1 describes that NI(T,Y)=1 presents a
necessary, but not sufficient, condition of an exact classification.

\textbf{Theorem 2}. For abstaining classification problems, when
$NI(T,Y)=0$, the classifier generally reflects a misclassification.
One special case is that all samples are considered to be one of $m$
classes, or be a reject class.

\textbf{Proof}. For NIs defined in Table 1, $NI(T,Y)=0$ \emph{iff}
$I(T,Y)=0$. According to information theory \cite{Cover:2006}, the
following conditions can hold based on the given marginal
distributions (or the empirical ones if a confusion matrix is used):

$$
I(T,Y) = 0,\quad iff \quad p(t,y) = p(t)p(y). \eqno{(10)}
$$
The conditional part in Eq. (10) can be rewritten in the form
$p_{ij}=p(t=t_i)p(y=y_j)$. From the constraints in (3), $p(t=t_i)>0
$ $ (i=1,2,\dots, m)$ can be obtained. For classification solutions,
there should exist at least one term for $p(y=y_j)>0 $ $
(j=1,2,\dots, m+1)$. Therefore, at least one non-zero term for
$p_{ij}>0 $ $ (i \ne j)$ must be obtained. This non-zero term
corresponds to the off-diagonal term in the confusion matrix, which
indicates that a misclassification has occurred. When all samples
have been identified as one of the classes (referring to M2 in Table
4, \cite{Hu:2008}), $NI=0$ should be obtained. $\diamondsuit$

\textbf{Remark 5}. Eq. (10) gives the statistical reason for zero
mutual information, that is, the two random variables are
``\emph{statistically independent}". Theorem 2 demonstrates an
intrinsic reason for local minima in NIs.

\textbf{Theorem 3}. The NI measures defined by the Shannon entropy
generally do not exhibit a monotonic property with respect to the
diagonal terms of a confusion matrix.

\textbf{Proof}. Based on \cite{Hu:2008}, we arrive at simpler
conditions for the local minima about $I(T,Y)$ for the given
confusion matrix:

$$
\textbf{C} = \left[ {\begin{array}{*{20}c}
   {\dots} & 0 & 0 & {\dots}  \\
   0 & {c_{i,i} } & {c_{i,i + 1} } & 0  \\
   0 & {c_{i + 1,i} } & {c_{i + 1,i + 1} } & 0  \\
   {\dots} & 0 & 0 & {\dots}  \\
\end{array}} \right],\quad
if \quad \frac{{c_{i,i} }}{{c_{i + 1,i} }} = \frac{{c_{i,i + 1}
}}{{c_{i + 1,i + 1} }}. \eqno{(11)}
$$

The local minima are obtained because the four given non-zero terms
in Eq. (11) produce zero (or the minimum) contribution to $I(T,Y)$.
Suppose a generic form is given for $NI(T,Y)=g(I(T,Y))$, where
$g(\cdotp)$ is a normalization function. From the chain rule of
derivatives, it can be seen that the conditions do not change for
reaching the local minima. $\diamondsuit$

\textbf{Remark 6}. The non-monotonic property of the information
measures implies that these measures may suffer from an intrinsic
problem of local minima for classification rankings (referring to
M19-M20 in Table 4, \cite{Hu:2008}). Or, according to Feature 1 of
the meta-measures, a rational result for the classification
evaluations may not be obtained due to the non-monotonic property of
the measures. This shortcoming has not been theoretically derived in previous
studies (\cite{Finn:1993}\cite{Forbes:1995}\cite{Tan:2004}).

\section{Normalized Information Measures based on Information
Divergence}

In this section, we propose normalized information measures based on
the definition of information divergence. In Table 2, we summarize
the commonly-used divergence measures, which are denoted as
$D_k(T,Y)$ and represents dissimilarity between the two random
variables $T$ and $Y$. In Sections 5 and 6, we apply the following
notations for defining marginal distributions:

$$
p_t (z) = p_t (t = z) = p(t), \quad and \quad p_y (z)  = p_y (y = z)
= p(y), \eqno{(12)}
$$
where $z$ is a possible scalar value that $t$ or $y$ can take. For a
consistent comparison with the previous normalized information
measures, we adopt the following transformation on $D_k$
\cite{Nishii:1999}:

$$
NI_k  = \exp ( - D_k ). \eqno{(13)}
$$

This transformation provides both inverse and normalization
functionalities. It does not introduce any extra parameters, and
presents a high degree of simplicity, as in derivation for examining
the local minima. Two more theorems are derived by following a
similar analysis to that in the previous section.

\textbf{Theorem 4}. For all NI measures in Table 2, when
$NI(T,Y)=1$, the classifier corresponds to the case of either an
exact classification, or a specific misclassification. Generally,
the misclassification in the latter case can not be removed by
switching labels in the confusion matrix.

\textbf{Proof}. When $p_y(z) = p_t(z)$, it is always the case that
$NI(T,Y)=1$. However, general conditions can be given for $p_y(z) =
p_t(z)$ as follows:

$$
p_y (y  = z_i ) = p_t (t  = z_i ),\quad or \quad \sum\limits_j
{p_{ji} } = \sum\limits_j {p_{ij} } , \quad i = 1,2,\dots, m
.\eqno{(14)}
$$
Eq. (14) implies two cases of classifications for $D_k(T,Y)=0,
k=10,\dots,20$, One of these corresponds to an exact classification
(or $y_k=t_k$), while the other is the result of a specific
misclassification that shows the relationship of $y_k \neq t_k$, but
$p_y(z) = p_t(z)$. In the latter case, switching labels in the
confusion matrix to remove misclassification generally destroys the
relation for $p_y(z) = p_t(z)$ at the same time. Considering the
relation is a necessary condition for a perfect classification, the
misclassification cannot be removed through a label switching
operation.
 $\diamondsuit$

\textbf{Remark 7}. Theorem 4 suggests the caution should be applied
in explaining the classification evaluations when $NI(T,Y)=1$. The
maximum of the NIs from the information divergence measures only
indicates the equivalence between the marginal probabilities,
$p_y(z) = p_t(z)$, but this is not always true for representing
exact classifications (or $y_k=t_k$). Theorem 4 reveals an intrinsic
problem when using an NI as a measure for similarity evaluations
between two datasets, such as in image registration.

\textbf{Theorem 5}. The NI measures based on information divergence
generally do not exhibit a monotonic property with respect to the
diagonal terms of confusion matrix.

\textbf{Proof}. The theorem can be proved by examining the existence
of multiple maxima for NI measures based on information divergence.
Here we use a binary classification as an example. The local minima
of $D_k$ are obtained when the following conditions exist for a
confusion matrix:

$$
\textbf{C} = \left[\begin{array}{*{20}c}
   {C_1  - d_1 } & {d_1 } & 0  \\
   {d_2 } & {C_2  - d_2 } & 0  \\
\end{array} \right]  \quad  and \quad d_1=d_2 ,
\eqno{(15)}
$$
where $d_1$ and $d_2$ are integer numbers ($>0$) for misclassified
samples. The confusion matrix in Eq. (15) produces zero divergence
$D_k$ and therefore, $NI_k=1$. However, changing from $d_1 \ne d_2$
always results in $NI_k<1$. $\diamondsuit$

\textbf{Remark 8}. Theorem 5 indicates another shortcoming of NIs in
the information divergence group from the viewpoint of monotonicity.
The reason is once again attributed to the usage of marginal
distributions in calculations of divergence. The shortcoming has not
been reported in previous investigations
(\cite{Nishii:1999}\cite{Li:2004}).

\section{Normalized Information Measures based on Cross-Entropy}
In this section, we propose normalized information measures based on
cross-entropy, which is defined for discrete random variables as
\cite{Bishop:1995}:

$$
H(T;Y) =  - \sum\limits_z {p_t (z)\log _2 p_y (z), \quad or \quad
H(Y;T) = - \sum\limits_z {p_y (z)\log _2 p_t (z)} }. \eqno{(16)}
$$
Note that $H(T;Y)$ differs from joint-entropy $H(T,Y)$ with respect
to both notation and definition, and is given as \cite{Cover:2006}:

$$
H(T,Y) =  - \sum\limits_t {\sum\limits_y {p(t,y)\log _2 } } p(t,y).
\eqno{(17)}
$$
In fact, from Eq. (16), one can derive the relation between KL
divergence (see Table 2) and cross-entropy:

$$
H(T;Y) = H(T) + KL(T,Y), \quad or \quad H(Y;T) = H(Y) + KL(Y,T).
\eqno{(18)}
$$

If $H(T)$ is considered as a constant in classification since the
target dataset is generally known and fixed, we can observe from Eq.
(18) that cross-entropy shares a similar meaning as KL divergence
for representing dissimilarity between $T$ and $Y$. From the
conditions $H \ge 0$ and $KL \ge 0$, we are able to realize the
normalization for cross-entropy shown in Table 3. Following similar
discussions as in the previous section, we can derive that all
information measures listed in Table 3 will also satisfy Theorems 4
and 5.

\section{Numerical Examples and Discussions}
This section presents several numerical examples together with
associated discussions. All calculations for the numerical examples
were done using the open source software Scilab
\footnote{\url{http://www.scilab.org}} and a specific toolbox
\footnote{ The toolbox is freely available as the file
``\emph{confmatrix2ni.zip}" at (\url{http://www.openpr.org.cn}).}.
The detailed implementation of this toolbox is described in
\cite{Hu:2009}. Table 4 lists six numerical examples in binary
classification problems according to the specific scenarios of their
confusion matrices. We adopt the notations from \cite{Chow:1970} for
the terms ``\emph{correct recognition rate (CR)}", ``\emph{error
rate (E)}", and ``\emph{reject rate (Rej)}" and their relation:

$$
CR+E+Rej=1. \eqno{(19)}
$$
In addition, we define ``\emph{accuracy rate (A)}" as

$$
A = \frac{{CR}}{{CR + E}} . \eqno{(20)}
$$
The first four classifications (or models) M1-M4 are provided to show
the specific differences with respect to error types and reject
types. In this work, we do not concern classifiers applied
(say, neural networks or support vector machines) for evaluations,
but only the resulting evaluations from these classifiers. In real
applications, it is common to encounter ranking classification
results as in M1 to M4. The first two classifications of M1 and M2
share the same values for the correct recognition and accuracy rates
($CR=A=99\%$). The other two classifications, for M3 and M4, have
the same reject rates ($Rej=1\%$) and correct recognition rates
($CR=99\%$). The accuracy rates for M3 and M4 are also the same
($A=100\%$). This definition is consistent with the conventions in
the study of ``\emph{Accuracy-Reject}" curves
\cite{Mackay:2003}. If neglecting the specific application backgrounds, 
users generally have a ranking order for the four classifications so that
the ``\emph{best}" one is selected. The data from other conventional
measures, such as $Precision$, $Recall$ and $F1$,
are also given in Table 4. Without using extra knowledge about the cost of
different error types or reject types, the conventional performance measures are not possible to rank
the four classifications, M1-M4, properly. 

According to the intuitions of Feature 3 proposed in Section 3, one
can gain two sets of ranking orders for the four classifications M1 to M4
in forms of:
$$
\Re (M2) > \Re (M1), \quad \Re (M4) > \Re (M3),  \eqno{(21-a)}
$$
$$
\Re (M4) > \Re (M2), \quad \Re (M3) > \Re (M1),  \eqno{(21-b)}
$$
where we denote $\Re (\bullet)$ to be a ranking operator, so that
$\Re (M_i)> \Re (M_j)$
expresses $M_i$ is better than $M_j$ in ranking. From eq. (21),
one is unable to tell the ranking order between M2 and M3.
For a fast comparison, a specific letter is assigned to the ranking order of
each model in Table 4 based on eq. (21):
$$
\Re (M4)=A, \Re (M3)=B,   \Re (M2)=B , \Re (M1)=C. \eqno{(22)}
$$
The top rank ``A" indicates the ``\emph{best}" classification (M4 in this
case) of the four models. Table 4 does not distinguish ranking
order between M2 and M3. However, numerical investigations using
information measures will provide
the ranking order from the given data. The other two models, M5 and M6, are
also specifically designed for the purpose of examining information
measures on Theorems 3 and 5 (or Feature 1), respectively.

Tables 5 and 6 present the results on information measures for
M1-M6, where the ranking orders among M1-M4 is based on the
calculation results of NIs with the given digits. The following
observations are achieved from the solutions to the examples.
\begin{itemize}
\item[1)] When normalization functions include the term $H(Y)$
for the mutual information group, the associated NI produces the
desirable feature of a variation in reject rate. $NI_2$ is effective
for this feature even if it only uses $H(T)$ for its normalization.
The effectiveness is attributed to the definition of $I_M(T,Y)$ for
calculating mutual information based on the intersection of $T$ and
$Y$.
\item[2)] The results of M5 and M6 confirm, respectively, Theorem 3 for
local minima and Theorem 5 for maxima of NIs. The existence of multi
extrema indicates the non-monotonic property with respect to the
diagonal terms of the confusion matrix, thereby exhibiting an
intrinsic shortcoming of the information measures.
\item[3)] For classifications M1 to M4, the meta-measure of Feature 3
suggests ranking orders as shown in eqs. (21) or (22). However, of all the
measures in the three groups only $NI_2$ shows any consistency with
the intuitions from the given examples (Tables 5 and 6). This result
indicates that Feature 3 seems to be a difficult property for most
information measures.
\item[4)] None of the performance or information measures investigated in this work fully
satisfy the meta-measures. Examining data distinguishability in M1
through M4, we consider the information measures from the
mutual-information group to be more appropriate than those of the
other groups (say, $NI_{12}$ and $NI_{22}$ do not show significant
distinguishability, or value differences, to the four models).
\end{itemize}

The fourth observation supports the proposal of
meta-measures for a higher level of classification evaluations. The
meta-measures provide users with a simple guideline of selecting
``\emph{proper}" measures from their specific concerns of applications. For example, 
the performance measures (such as $A$, $E$, $CR$, $F1$, or AUC) satisfy 
Feature 1, but fail directly to distinguish error types and reject types in
an objective evaluation. When Feature 2 or 3 is a main concern, the information measures 
exhibited to be more effective, despite them not being perfect. 

Of all the information measures investigated, $NI_2$ is shown to be the
``\emph{best}" for the given examples in terms of Feature 3.
Therefore, more detailed studies, from both theoretical and
numerical ones, were made on this promising measure.
The theoretical properties of this measure was derived in Appendix A.
While Theorem A1 confirms that $NI_2$ satisfy Feature 3 around the
exact classifications, Theorem A2 indicates that this measure
is able to adjust the ranking order between a misclassification
of a large class and a rejection of a small class.
Table 7 shows two sets of confusions matrices which are similar to M1-M4
in Table 4. One can observe the changes of ranking orders among them.
These changes numerically confirm Theorem A2 and its critical point,
or cross-over point ($\Omega =C_1/n \approx 0.942$), for the given data.

Further investigations were carried out on three-class problems.
Although some NIs could be removed directly based on their poor
performance with respect to the meta-measures (such as $NI_1$ and
$NI_9$ on Feature 2), they were retained to demonstrate pros and
cons in the applications. At this stage, we extend the concepts of
error types and reject types to multiple classes. Nine examples are
specifically designed in Table 8. The ranking order for each model is 
shown in Table 8, which is derived from the intuitions of Feature 3.
From Tables 9 and 10, it
is interesting to see that $NI_2$ is still the most appropriate
measure for classification evaluations. Using this measure, we can
select the ``\emph{best}" and ``\emph{worst}" classifications
consistent with our intuition. All other measures perform below our
satisfactions for distinguishing error types and reject types properly.

The numerical study supports the viewpoint that no universally superior measure exists.
For example, in comparing with information measure $NI_2$,
the conventional accuracy measure satisfies
Feature 1, but does not qualify to Feature 3.
Thus, any measure, either performance-based or information-based, should
be designed and evaluated within the context of the specific
applications. It is evident that the desirable features in the
specific applications become more crucial (or ``\emph{proper}") for
evaluation measures than some generic mathematical properties. For
example, information measures (such as KL divergence), that may not
satisfy a metric's properties (say, symmetry), are able to process
classification evaluations including a reject option. They provide
more applicable power than the conventional performance measures in
abstaining classifications. However, we still need a complete
picture about information measures with respect to their advantages as well
as limitations. The examples in Tables 4, 7, and 8 only present limited
scenarios for variations in confusion matrices. Using the open-source
toolbox from \cite{Hu:2009}, one is able to test more scenarios
for numerical investigations.

\section{Summary}
In this work, we investigated objective evaluations of
classifications by introducing normalized information measures.
We reviewed the related works and discussed objectivity and its
formal definition in evaluations. Objective evaluations may be
required under different application background.
In classifications, for example, exact knowledge of misclassification
costs is sometimes unknown in evaluations.
Moreover, cases of ignorance regarding reject costs
appear more often in scenarios of abstaining classifications. In
these cases, although subjective evaluations can be applied, the
user-given data of the unknown abstention costs will lead to a much
higher degree of uncertainty or inconsistency. We believe that an
objective evaluation can be a suitable solution, as well as a complementary, approach to
subjective evaluations. In some situations, an objective evaluation
is considered useful despite the subjective evaluations being
reasonable for the applications. The results from both objective and
subjective evaluations give users an overall quality of
classification results.

Considering that abstaining classifications are becoming more
popular, we focused on the distinctions of error types and reject
types within objective evaluations of classifications. First, we
proposed three meta-measures for assessing classifications, which
seem more relevant and proper than the properties of metrics in the
context of classification applications. The meta-measures provide
users with useful guidelines for a quick selection of candidate
measures. Second, we tried systematically to enrich a classification
evaluation bank by including commonly used information measures.
Contrary to the conventional performance measures that apply
empirical formulas, the information measures are theoretically more
sound for objective evaluations of classifications. The key
advantage of these measures over the conventional ones is their
ability to handle multi-class classification evaluations with a
reject option. Third, we revealed theoretically the intrinsic
shortcomings of the information measures. These have not been
formally reported before in studies of image registration, feature selection,
or similarity ranking. The discovery of these shortcomings is very
important for users to interpret their results correctly when
applying those measures.

Based on the principle of the ``\emph{No Free-Lunch Theorem}"
\cite{Duda:2001}, we recognize that there are no ``\emph{universally
superior}" measures \cite{Lavesson:2007}. It is not our aim to
replace the conventional performance measures, but to explore
information measures systematically in classification evaluations.
The theoretical study demonstrates the strength and weakness of the
information measures. Numerical investigations, conducted on binary
and three-class classifications, confirmed that objective
evaluations are not an easy topic in the study of machine learning.
One of the most challenging tasks will be an exploration of novel
measures that satisfy all meta-measures as well as
the metric properties in objective evaluations of classifications.
It is also necessary to define the ``\emph{ranking order}" intuitions
among error types and reject types in
generic classifications, which will form the basis of the quantitative
meta-measures. However, this task becomes more difficult if
classifications are beyond two classes.


\section*{Acknowledgment}
This work is supported in part by Natural Science of Foundation of
China (\#61075051).

\appendix
\section*{Appendix A. Theorems and Sensitivity Functions of $NI_2$ for
Binary Classifications}

\textbf{Theorem A1}: For a binary classification defined by:
$$
\textbf{C} = \left[ {\begin{array}{*{20}c}
   {TN} & {FP} & {RN}  \\
   {FN} & {TP} & {RP}  \\
\end{array}} \right], \quad \rm{and}
\eqno{({\rm{A1-a}})}
$$
$$
{\rm{C}}_{\rm{1}}  = TN + FP + RN, C_2  = FN + TP + RP, C_1  + C_2 =
n \eqno{({\rm{A1-b}})}
$$
$NI_2$ satisfies Feature 3 on the property regarding error types and
reject types
 around the exact classifications. Specifically for the
four confusion matrices below:
$$
\begin{array}{*{20}c}
   {M_1  = \left[ {\begin{array}{*{20}c}
   {C_1 } & 0 & 0  \\
   d & {C_2  - d} & 0  \\
\end{array}} \right],} & {M_2  = \left[ {\begin{array}{*{20}c}
   {C_1  - d} & d & 0  \\
   0 & {C_2 } & 0  \\
\end{array}} \right],}  \\
   {M_3  = \left[ {\begin{array}{*{20}c}
   {C_1 } & 0 & 0  \\
   0 & {C_2  - d} & d  \\
\end{array}} \right],} & {M_4  = \left[ {\begin{array}{*{20}c}
   {C_1  - d} & 0 & d  \\
   0 & {C_2 } & 0  \\
\end{array}} \right],}  \\
\end{array} \eqno{({\rm{A2}})}
$$
the following relations will be held:
$$
NI_2 (M_1 ) < NI_2 (M_2 ) \quad and \quad NI_2 (M_3 ) <
NI_2 (M_4 ) ,
\eqno{({\rm{A3-a}})}
$$
$$
NI_2 (M_1 ) < NI_2 (M_3 ) \quad and \quad NI_2 (M_2 ) < NI_2 (M_4 ),
\eqno{({\rm{A3-b}})}
$$
$$
\quad where  \quad  C_1  > C_2  > d >0.
\eqno{({\rm{A3-c}})}
$$

\textbf{Proof}. For a binary classification, $NI_2$ is defined by the
modified mutual information:
$$
\begin{array}{l}
 NI_2  = \frac{{I_M (T,Y)}}{{H(T)}},\quad and \\
 I_M (T,Y) = \frac{{TN}}{n}\log _2 \frac{{nTN}}{{C_1 (TN + FN)}} + \frac{{FP}}{n}\log _2 \frac{{nFP}}{{C_1 (TP + FP)}} \\
 \begin{array}{*{20}c}
   {} & {} & { + \frac{{FN}}{n}\log _2 }  \\
\end{array}\frac{{nFN}}{{C_2 (FN + TN)}} + \frac{{TP}}{n}\log _2 \frac{{nTP}}{{C_2 (FP + TP)}}. \\
 \end{array}
  \eqno{({\rm{A4}})}
$$
Let $M_0$ be a confusion matrix corresponding to the exact
classifications:
$$
M_0  = \left[ {\begin{array}{*{20}c}
   {C_1 } & 0 & 0  \\
   0 & {C_2 } & 0  \\
\end{array}} \right]. \eqno{({\rm{A5}})}
$$
Based on the definition of $I_M$ in (A4), one can calculate the
mutual information differences between two models. Considering $M_0$
to be a baseline, we obtain the analytical results below for the
four models:
$$
\Delta I_{10}  = I_M(M_1 ) - I_M(M_0 ) = \frac{1}{n}(C_1 \log _2
\frac{{C_1 }}{{C_1  + d}} + d\log _2 \frac{d}{{C_1  + d}}),
\eqno{({\rm{A6-a}})}
$$
$$
\Delta I_{20}  = I_M(M_2 ) - I_M(M_0 ) = \frac{1}{n}(C_2 \log _2
\frac{{C_2 }}{{C_2  + d}} + d\log _2 \frac{d}{{C_2  + d}}),
\eqno{({\rm{A6-b}})}
$$
$$
\Delta I_{30}  = I_M(M_3 ) - I_M(M_0 ) = \frac{d}{n}(\log _2 \frac{{C_2
}}{n}), \eqno{({\rm{A6-c}})}
$$
$$
\Delta I_{40}  = I_M(M_4 ) - I_M(M_0 ) = \frac{d}{n}(\log _2 \frac{{C_1
}}{n}), \eqno{({\rm{A6-d}})}
$$
For the given assumption $C_1>C_2>d>0$, all $\Delta I$s  above are
negative values so that their abstracts represent the absolute costs
in classifications. One can directly prove that $|\Delta
I_{30}|>|\Delta I_{40}|$ from (A6-c) and (A6-d). The procedures for
the proof of $|\Delta I_{10}|>|\Delta I_{20}|$ are given below.
First, one needs to confirm the following two functions to be
strictly decreasing $(x_1 < x_2, g(x_1)>g(x_2))$:
$$
g_1 (x) = (\frac{x}{{x + d}})^x \quad and \quad g_2 (x) =
(\frac{d}{{x + d}})^d \quad for \quad x > 0, \; d > 0.
\eqno{({\rm{A7-a}})}
$$
Then, from the monotonically decreasing property of (A7-a), one can
derive the following relations:
$$
\begin{array}{l}
 C_1  > C_2  \to (\frac{{C_1 }}{{C_1  + d}})^{C_1 }  < (\frac{{C_2 }}{{C_2  + d}})^{C_2 }  < 1 \quad and \quad (\frac{d}{{C_1  + d}})^d  < (\frac{d}{{C_2  + d}})^d  < 1 \\
  \to \frac{1}{n}|C_2 \log _2 \frac{{C_2 }}{{C_2  + d}} + d\log _2 \frac{d}{{C_2  + d}}| < \frac{1}{n}|C_1 \log _2 \frac{{C_1 }}{{C_1  + d}} + d\log _2 \frac{d}{{C_1  + d}}| \\
  \to |\Delta I_{20} | < |\Delta I_{10} | \\
 \end{array}
 \eqno{({\rm{A7-b}})}
$$
The relations in (A3-a) are achieved for $NI_2$ because its
normalization term, $H(T)$, is a constant for the given $C_1$ and
$C_2$. One therefore confirms the satisfaction of Feature 3 on the
property of the within error types and reject types around the exact
classifications, respectively.

Then it is a proof of the relation (A3-b), which suggests that a misclassification suffer
a higher cost than a rejection for the same class. Feature 3
considers this relation as a basic property in classifications for
the between error and reject types. The procedures for the proof
are:
$$
\begin{array}{l}
 C_1  > C_2  \to C_1 C_2  + C_1 d > (C_1  + C_2 )d = nd \\
  \to 1 > \frac{{C_1 }}{n} > \frac{d}{{C_2  + d}} \to \left| {\log _2 (\frac{{C_1 }}{n})} \right| < \left| {\log _2 (\frac{d}{{C_2  + d}})} \right| \\
  \to \frac{1}{n}\left| {d\log _2 (\frac{{C_1 }}{n})} \right| < \frac{1}{n}\left| {d\log _2 \frac{d}{{C_2  + d}}} \right| < \frac{1}{n}\left| {C_2 \log _2 \frac{{C_2 }}{{C_2  + d}} + d\log _2 \frac{d}{{C_2  + d}}}
  \right|\\
  \to \left| {\Delta I_{40} } \right| < \left| {\Delta I_{20} } \right| \\
 \end{array}
 \eqno{({\rm{A8-a}})}
$$
$$
\begin{array}{l}
 C_1  + d < n \to C_1 (C_1  + d) + nd < C_1 n + nd \to \frac{{C_1 (C_1  + d) + nd}}{{n(C_1  + d)}} < 1 \\
  \to \frac{{C_1 }}{n} + \frac{d}{{C_1  + d}} < 1 \to \frac{d}{{C_1  + d}} < \frac{{C_2 }}{n} < 1 \to \left| {\log _2 \frac{{C_2 }}{n}} \right| < \left| {\log _2 \frac{d}{{C_1  + d}}} \right| \\
  \to \frac{1}{n}\left| {d\log _2 \frac{{C_2 }}{n}} \right| < \frac{1}{n}\left| {C_1 \log _2 \frac{{C_1 }}{{C_1  + d}} + d\log _2 \frac{d}{{C_1  + d}}} \right| \\
  \to \left| {\Delta I_{30} } \right| < \left| {\Delta I_{10} } \right|. \\
 \end{array}
 \eqno{({\rm{A8-b}})}
$$
$ \diamondsuit$

\textbf{Theorem A2}: For the given conditions (A1)-(A2) and $C_1>C_2>d>0$,
$NI_2$ will satisfy the following relations:
$$
NI_2 (M_4 ) > NI_2 (M_3 ) > NI_2 (M_2 ) > NI_2 (M_1) \quad for \quad
0.5 < p_1 < \Omega  \le 1
 \eqno{({\rm{A9-a}})}
$$
$$
NI_2 (M_4 ) > NI_2 (M_2 ) > NI_2 (M_3 ) > NI_2 (M_1 ) \quad for
\quad 0.5 < \Omega < p_1  \le 1
 \eqno{({\rm{A9-b}})}
$$
where we set $p_1=C_1/n$, and $\Omega$ is an upper boundary for the
validation of (A9-a).

\textbf{Proof}. The first relation describes that
the ranking order in (A9-a) is valid only for a certain range of
$p_1$. The lower boundary is resulted from the assumption of $C_1$
to be a large class. The upper boundary, $\Omega$, is determined
by the cross-over point between the functions of (A-6b) and (A-6c).
For better understanding of the relations (A9), we present the
plots of ``$\Delta I$ vs. $p_1$'' when $n=100$ and $d=1$
(Fig. A1).

\begin{figure}[htb]
\renewcommand{\thefigure}{A\arabic{figure}.}
    \centering
    \includegraphics[width=110mm]{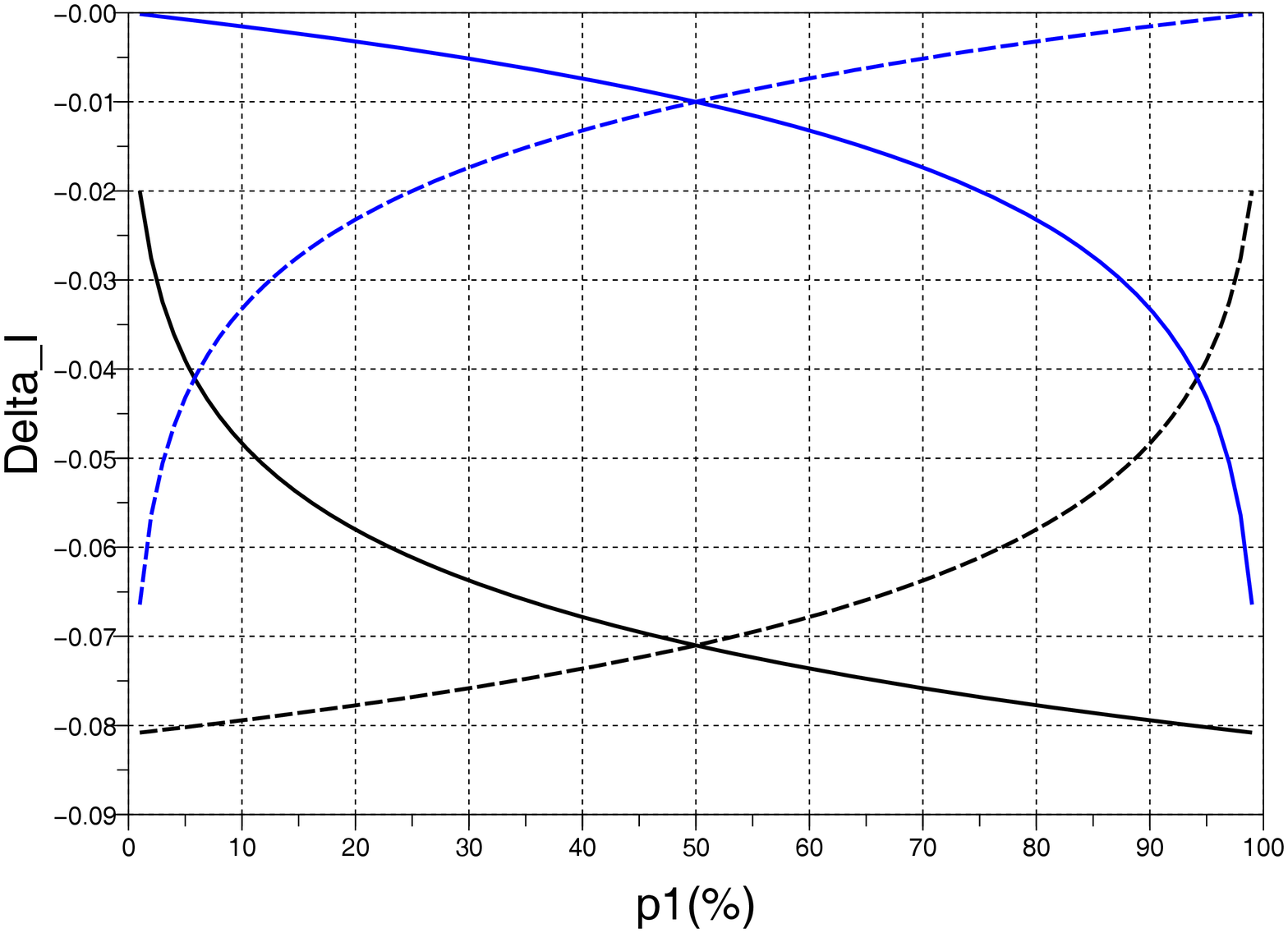}
    \caption{Plots of ``$\Delta I$ vs. $p_1(\%)$'' when $n=100$ and $d=1$. \newline
(Black-Solid = $\Delta I_{10}$, Black-Dash=$\Delta I_{20}$,
Blue-Solid=$\Delta I_{30}$, Blue-Dash=$\Delta I_{40}$)}
\end{figure}

For examining the validation range of (A9-a), one needs to
calculate the cross-over point from solving the equation below:
$$
f = \Delta I_{20}  - \Delta I_{30}  = \frac{1}{n}(C_2 \log _2
\frac{{C_2 }}{{C_2  + d}} + d\log _2 \frac{{dn}}{{C_2  + d}}) = 0.
\eqno{({\rm{A10}})}
$$
There exists no closed-form solution to $\Omega$. Based on the
monotonicity of the related functions and relations in (A3),
one is able to confirm the conditions in (A9-a) and
(A9-b), respectively. Fig. A1 depicts numerically that only a
single cross-over point appears to the range of $p_1>0.5$(or
$C_1>C_2$). $ \diamondsuit$

\textbf{Remark A1}: We can denote $\Omega(n,d)$ to be the cross-over point
obtained from $f$, with two independent variables $n$ and $d$. The
value of $\Omega$ increases with $n$, but decreases with $d$. A
numerical solution to $\Omega$ should be engaged. The physical
interpretation of $\Omega$ is a critical point at which a rejection
within a small class has the same cost with a misclassification
within a large class. This situation generally does not occur
except for classifications of largely-skewed classes (say,
$C_1>>C_2$). Therefore, we call the ranking order in (A9-a) is a
general ranking order, and one in (A9-b) is a
largely-skewed-class ranking order.

\textbf{Sensitivity functions}: The sensitivity functions are given
as the conventional forms for delivering approximation analysis of
$I_M$:
$$
\frac{{\partial I_M}}{{\partial TN}} = \frac{1}{n}\left[ {\log _2
\frac{n}{{C_1 }} + \left( {\log _2 \frac{{TN}}{{TN + FN}}}
\right)sng(TN)} \right], \eqno{({\rm{A11-a}})}
$$
$$
\frac{{\partial I_M}}{{\partial TP}} = \frac{1}{n}\left[ {\log _2
\frac{n}{{C_2 }} + \left( {\log _2 \frac{{TP}}{{TP + FP}}}
\right)sng(TP)} \right], \eqno{({\rm{A11-b}})}
$$
$$
\frac{{\partial I_M}}{{\partial FN}} = \frac{1}{n}\left[ {\log _2
\frac{n}{{C_2 }} + \left( {\log _2 \frac{{FN}}{{FN + TN}}}
\right)sng(FN)} \right], \eqno{({\rm{A11-c}})}
$$
$$
\frac{{\partial I_M}}{{\partial FP}} = \frac{1}{n}\left[ {\log _2
\frac{n}{{C_1 }} + \left( {\log _2 \frac{{FP}}{{FP + TP}}}
\right)sng(FP)} \right], \eqno{({\rm{A11-d}})}
$$
$$
\frac{{\partial I_M}}{{\partial RN}} =  - \frac{{\partial
I}}{{\partial TN}} - \frac{{\partial I}}{{\partial FP}},
\eqno{({\rm{A11-e}})}
$$
$$
\frac{{\partial I_M}}{{\partial RP}} =  - \frac{{\partial
I}}{{\partial FN}} - \frac{{\partial I}}{{\partial TP}}.
\eqno{({\rm{A11-f}})}
$$
where $sgn(.)$ is a sign function for satisfying the definition of
$H(0)=0$. Only four independent variables describe the sensitivity
functions due to the two constraints in (A1-b). Hence, a chain rule
is applied for deriving the functions of (A11-e) and (A11-f).
$\diamondsuit$

\textbf{Remark A2}: Using eq.(A11), we failed to reach the reasonable
conclusions as those in Theorems A1 for the reason that the
first-order differentials may be not sufficient for the analysis
around the exact classifications. For example, we got the results
for:
$$
\begin{array}{l}
 I(M_1 ) - I(M_0 ) \approx (TP_1  - TP_0 )\frac{{\partial I_M(M_0 )}}{{\partial TP}} + (FN_1  - FN_0 )\frac{{\partial I_M(M_0 )}}{{\partial FN}} \\
 \begin{array}{*{20}c}
   {} & {} & { =  - \frac{d}{n}\log _2 (\frac{n}{{C_2 }}) + \frac{d}{n}\log _2 (\frac{n}{{C_2 }}) = 0.}  \\
\end{array} \\
 \end{array}
 \eqno{({\rm{A12-a}})}
$$
$$
\begin{array}{l}
 I(M_2 ) - I(M_0 ) \approx (TN_1  - TN_0 )\frac{{\partial I_M(M_0 )}}{{\partial TN}} + (FP_1  - FP_0 )\frac{{\partial I_M(M_0 )}}{{\partial FP}} \\
 \begin{array}{*{20}c}
   {} & {} & { =  - \frac{d}{n}\log _2 (\frac{n}{{C_1 }}) + \frac{d}{n}\log _2 (\frac{n}{{C_1 }}) = 0.}  \\
\end{array} \\
 \end{array} \eqno{({\rm{A12-b}})}
$$
This observation suggests that one needs to be cautious when using
sensitivity function for approximation analysis on $I_M$ (or $NI_2$).



\bibliographystyle{elsarticle-num}
\bibliography{reference1}

\newpage

\begin{table*}[htb]
\addtolength{\tabcolsep}{-3.5pt}
\begin{center}
\caption{NI measures within the mutual-information based group.}
\small
\begin{tabular}{lll}
    \hline\noalign{\smallskip}
    No. & Name [Reference] & Formula on $NI_k$\\
    \hline\noalign{\smallskip}
    1 & NI based on mutual information \cite{Finn:1993} & $ NI_1 (T,Y) = \frac{{I(T,Y)}}{{H(T)}}$\\
    2 & NI based on mutual information \cite{Hu:2008}   & $NI_2 (T,Y) = \frac{{I_M (T,Y)}}{{H(T)}} $\\
    3 & NI based on mutual information \cite{Finn:1993} & $ NI_3 (T,Y) = \frac{{I(T,Y)}}{{H(Y)}}$\\
    4 & NI based on mutual information & $NI_4 (T,Y) = \frac{1}{2}\left[ {\frac{{I(T,Y)}}{{H(T)}} +
\frac{{I(T,Y)}}{{H(Y)}}} \right] $\\
    5 & NI based on mutual information \cite{Kvalseth:1987} & $NI_5 (T,Y) = \frac{{2I(T,Y)}}{{H(T) + H(Y)}} $\\
    6 & NI based on mutual information \cite{Strehl:2002} & $NI_6 (T,Y) = \frac{{I(T,Y)}}{{\sqrt {H(T)H(Y)} }} $\\
    7 & NI based on mutual information \cite{Malvestuto:1986} & $NI_7 (T,Y) = \frac{{I(T,Y)}}{{H(T,Y)}} $\\
    8 & NI based on mutual information \cite{Kvalseth:1987}& $NI_8 (T,Y) = \frac{{I(T,Y)}}{{\max (H(T),H(Y))}} $ \\
    9 & NI based on mutual information \cite{Kvalseth:1987}& $NI_9 (T,Y) = \frac{{I(T,Y)}}{{\min (H(T),H(Y))}} $ \\
    \noalign{\smallskip}\hline
\end{tabular}
\end{center}
\end{table*}

\begin{landscape}
\begin{table*}[htb]
\begin{center}
\caption{Information measures within the divergence based group.}
\small
\begin{tabular}{lll}
    \hline\noalign{\smallskip}
    No. & Name of $D_k$ [Reference] & Formula on $D_k$ ($NI_k=exp(-D_k)$)\\
    \hline\noalign{\smallskip}
    10 & ED-Quadratic Divergence \cite{Principe:2000} & $D_{10}  = QD_{ED} (T,Y) = \sum\limits_z {(p_t (z) - p_y (z))^2 } $\\
    11 & CS-Quadratic Divergence \cite{Principe:2000} & $D_{11}  = QD_{CS} (T,Y) = \log _2 \frac{{\sum\limits_z {p_t (z)^2 }
\sum\limits_z {p_y (z)^2 } }}{{[\sum\limits_z {(p_t (z)} p_y (z))]^2}} $\\
    12 & KL Divergence \cite{Kullback:1951} & $D_{12}  = KL(T,Y) = \sum\limits_z {p_t (z)\log _2 \frac{{p_t
(z)}}{{p_y (z)}}} $\\
    13 & Bhattacharyya Distance \cite{Johnson:2001} & $D_{13}  = D_B (T,Y) =  - \log _2 \sum\limits_z {\sqrt {p_t (z)p_y
(z)} } $\\
    14 & $\chi ^2 $ (Pearson) Divergence \cite{Csiszar:1963}& $D_{14}  = \chi ^2 (T,Y) = \sum\limits_z {\frac{{(p_t (z) - p_y
(z))^2 }}{{p_y (z)}}} $\\
    15 & Hellinger Distance \cite{Csiszar:1963}& $D_{15}  = H^2 (T,Y) = \sum\limits_z {(\sqrt {p_t (z)}  - \sqrt {p_y
(z)} )^2 } $\\
    16 & Variation Distance \cite{Csiszar:1963}& $D_{16}  = V(T,Y) = \sum\limits_z {|p_t (z) - p_y (z)|} $\\
    17 & J divergence  j \cite{Lin:1991}&
    $D_{17}  = J(T,Y) = \sum\limits_z {p_t (z)\log _2 \frac{{p_t
(z)}}{{p_y (z)}}}  + \sum\limits_z {p_y (z)\log _2 \frac{{p_y
(z)}}{{p_t (z)}}} $\\
    18 & L (or JS) divergence \cite{Lin:1991}& $D_{18}  = L(T,Y) = KL(T,M) + KL(Y,M),M = \frac{{(p_t (z) + p_y
(z))}}{2} $\\
    19 & Symmetric $\chi ^2 $ Divergence \cite{Malerba:2002} &
    $D_{19}  = \chi _S^2 (T,Y) = \sum\limits_z {\frac{{(p_t (z) - p_y
(z))^2 }}{{p_y (z)}}}  + \sum\limits_z {\frac{{(p_y (z) - p_t (z))^2
}}{{p_t (z)}}} $\\
    20 & Resistor Average Distance \cite{Johnson:2001} &
    $D_{20}  = D_{RA} (T,Y) = \frac{{KL(T,Y)KL(Y,T)}}{{KL(T,Y) +
KL(Y,T)}}$
\\
    \noalign{\smallskip}\hline
\end{tabular}
\end{center}
\end{table*}
\end{landscape}

\begin{table*}[htb]
\begin{center}
\caption{NI measures within the cross-entropy based group.} \small
\begin{tabular}{lll}
    \hline\noalign{\smallskip}
    No. & Name  & Formula on $NI_k$\\
    \hline\noalign{\smallskip}
    21 & NI based on cross-entropy & $NI_{2{\rm{1}}}  = \frac{{H(T)}}{{H(T;Y)}},H(T;Y) =  - \sum\limits_z
{p_t (z)\log _2 p_y (z)}$\\
    22 & NI based on cross-entropy & $NI_{22}  = \frac{{H(Y)}}{{H(Y;T)}},H(Y;T) =  - \sum\limits_z {p_y
(z)\log _2 p_t (z)}$\\
    23 & NI based on cross-entropy & $NI_{23}  = \frac{1}{2} \left (\frac{{H(T)}}{{H(T;Y)}} +
\frac{{H(Y)}}{{H(Y;T)}} \right ) $\\
    24 & NI based on cross-entropy & $NI_{24}  = \frac{{H(T)+H(Y)}}{{H(T;Y)+H(Y;T)}} $\\
    \noalign{\smallskip}\hline
\end{tabular}
\end{center}
\end{table*}

\begin{table*}[htb]
\addtolength{\tabcolsep}{-5pt}
\begin{center}
\caption{Numerical examples in Binary Classifications(M1-M4 and M6:
$C_1=90,C_2=10$; M5: $C_1=95,C_2=5$). (R)= ranking order for the
model, where R = A,B, ..., in descending order from the top.} \small
\begin{tabular}{lllllll}
    \hline\noalign{\smallskip}
    Model  & M1  & M2  &M3 & M4  & M5 & M6\\
    (Ranking) & (C) & (B) & (B)& (A) \\
    \hline\noalign{\smallskip}
    \textbf{C} & $\left[ {\begin{array}{*{20}c}   90 & 0 & 0  \\   1 & 9 & 0  \\ \end{array}}
    \right]$ & $
\left[ {\begin{array}{*{20}c}
   89 & 1 & 0  \\
   0 & 10 & 0  \\
\end{array}} \right]$
& $ \left[ {\begin{array}{*{20}c}
   90 & 0 & 0  \\
   0 & 9 & 1  \\
\end{array}} \right]$
& $ \left[ {\begin{array}{*{20}c}
   89 & 0 & 1  \\
   0 & 10 & 0  \\
\end{array}} \right]$
& $ \left[ {\begin{array}{*{20}c}
   {57} & {38} & 0  \\
   3 & 2 & 0  \\
\end{array}} \right]$
& $ \left[ {\begin{array}{*{20}c}
   89 & 1 & 0  \\
   1 & 9 & 0  \\
\end{array}} \right]
$\\
\emph{CR} & 0.990 & 0.990 & 0.990 & 0.990 & 0.590 & 0.980 \\
\emph{Rej} & 0.000 & 0.000 & 0.010 & 0.010 & 0.000 & 0.000 \\
\emph{Precision} & 0.989 & 1.000 & 1.000 & 1.000 & 0.950 & 0.989 \\
\emph{Recall} & 1.000 & 0.989 & 1.000 & 1.000 & 0.600 & 0.989 \\
\emph{F1} & 0.994 & 0.994 & 1.000 & 1.000 & 0.735 & 0.989 \\
\noalign{\smallskip}\hline
\end{tabular}
\end{center}
\end{table*}

\begin{landscape}
\begin{table*}[htb]
\begin{center}
\caption{Results for the models in Table 4 on information measures
from mutual-information and cross-entropy groups. (R)= ranking order
for the model, where R = A,B, ..., in descending order from the top.
} \small
\begin{tabular}{llllllllllllll}
 \hline\noalign{\smallskip}
 Model & $NI_1$ & $NI_2$ & $NI_3$ & $NI_4$ & $NI_5$ & $NI_6$ & $NI_7$ & $NI_8$ & $NI_9$ & $NI_{22}$& $NI_{23}$& $NI_{24}$& $NI_{25}$\\
 \hline\noalign{\smallskip}
  M1 & 0.831 & 0.831 & 0.893 & 0.862 & 0.860 & 0.861 & 0.755 & 0.831 & 0.893 & 0.998 & 0.998 & 0.998 & 0.998\\
  (C)& (D) & (D) & (B) & (D) & (D) & (D) & (D) & (D) & (D) & (A) & (A) & (A) & (A)\\
  \hline
  M2 & 0.897 & 0.897 & 0.841 & 0.869 & 0.868 & 0.869 & 0.767 & 0.841 & 0.897 & 0.998 & 0.998 & 0.998 & 0.998\\
  (B)& (C) & (C) & (D) & (C) & (C) & (C) & (C) & (C) & (C) & (A) & (A) & (A) & (A)\\
  \hline
  M3 & 1.000 & 0.929 & 0.909 & 0.955 & 0.952 & 0.953 & 0.909 & 0.909 & 1.000 & 0.969 & 0.000 & 0.484 & 0.000\\
  (B) & (A) & (B) & (A) & (A) & (A) & (A) & (A) & (A) & (A) & (D) & (B) & (C) & (B)\\
  \hline
  M4 & 1.000 & 0.997 & 0.855 & 0.928 & 0.922 & 0.925 & 0.855& 0.855& 1.000& 0.970& 0.000& 0.485& 0.000\\
  (A) &  (A) &  (A) &  (C) & (B) & (B) & (B) & (B) & (B) & (A) & (C) & (B) & (B) & (B) \\
  \hline
  M5 & 0.000 & 0.000 & 0.000 & 0.000 & 0.000 & 0.000 & 0.000 & 0.000 & 0.000 & 0.374 & 0.548 & 0.461 & 0.495\\
 \hline
  M6 & 0.731 & 0.731 & 0.731 & 0.731 & 0.731 & 0.731 & 0.576 & 0.731 & 0.731 & 1.000 & 1.000 & 1.000 &
  1.000\\
  \noalign{\smallskip}\hline
\end{tabular}
\end{center}
\end{table*}
\end{landscape}

\begin{landscape}
\begin{table*}[htb]
\begin{center}
\caption{Results for the models in Table 4 on information measures
from divergence group. S=singularity which cannot be removed. (R)=
ranking order for the model, where R = A,B, ..., in descending order
from the top. } \small
\begin{tabular}{llllllllllll}
 \hline\noalign{\smallskip}
 Model & $NI_{10}$ & $NI_{11}$ & $NI_{12}$ & $NI_{13}$ & $NI_{14}$ & $NI_{15}$ & $NI_{16}$ & $NI_{17}$ & $NI_{18}$ & $NI_{19}$ & $NI_{20}$\\
 \hline\noalign{\smallskip}
  M1 & 0.9998 & 0.9998 & 0.9991 & 0.9998 & 0.9988 & 0.9997 & 0.9802 & 0.9983 & 0.9996 & 0.9977 & 0.9996\\
  (C) & (A) & (A) & (B) & (A) & (B) & (A) & (A) & (B) & (A) & (B) &  (A)\\
 \hline
  M2 & 0.9998 & 0.9998 & 0.9992 & 0.9998 & 0.9990 & 0.9997 & 0.9802 & 0.9985 & 0.9996 & 0.9979 & 0.9996 \\
  (B) & (A) & (A) & (A) & (A) & (A) & (A) & (A) & (A) & (A) & (A) & (A)\\
 \hline
  M3 & 0.9998 & 0.9996 & 0.9849 & 0.9926 & 0.9890 & 0.9898 & 0.9802 & S & 0.9897 & S & S\\
  (B) & (A) & (D) & (D) & (D) & (D) & (D) & (A)  &  &  (D) &  & \\
 \hline
  M4 & 0.9998 & 0.9998 & 0.9856 & 0.9928 & 0.9899 & 0.9900 & 0.9802 & S & 0.9900 & S & S \\
  (A) & (A) & (A) & (C) & (C) & (C) & (C) & (A) &  & (C) &  & \\
 \hline
  M5 & 0.7827 & 0.6473 & 0.6189 & 0.8540 & 0.6002 & 0.8129 & 0.4966 & 0.2775 & 0.7550 & 0.0455 & 0.7406 \\
 \hline
  M6 & 1.0000 & 1.0000 & 1.0000 & 1.0000 & 1.0000 & 1.0000 & 1.0000 & 1.0000 & 1.0000 & 1.0000 & S\\
 \noalign{\smallskip}\hline
\end{tabular}
\end{center}
\end{table*}
\end{landscape}

\begin{landscape}
\begin{table*}[htb]
\begin{center}
\caption{Numerical examples in Binary Classifications(n=100). (R)= ranking order for the model,
where R = A,B, ..., in descending order from the top.}
\small
\begin{tabular}{ccccccccc}
 \hline
 Model & M1a & M2a & M3a & M4a & M1b & M2b & M3b & M4b \\
 \hline
 C & $
\left[ {\begin{array}{*{20}c}
   {94} & 0 & 0  \\
   1 & 5 & 0  \\
\end{array}} \right]$ &
$ \left[ {\begin{array}{*{20}c}
   {93} & 1 & 0  \\
   0 & 6 & 0  \\
\end{array}} \right]$ &
$ \left[ {\begin{array}{*{20}c}
   {94} & 0 & 0  \\
   0 & 5 & 1  \\
\end{array}} \right]
$ &
$\left[ {\begin{array}{*{20}c}
   {93} & 0 & 1  \\
   0 & 6 & 0  \\
\end{array}} \right]$&
$\left[ {\begin{array}{*{20}c}
   {95} & 0 & 0  \\
   1 & 4 & 0  \\
\end{array}} \right]$&
$\left[ {\begin{array}{*{20}c}
   {94} & 1 & 0  \\
   0 & 5 & 0  \\
\end{array}} \right]$ &
$\left[ {\begin{array}{*{20}c}
   {95} & 0 & 0  \\
   0 & 4 & 1  \\
\end{array}} \right]$ &
$\left[ {\begin{array}{*{20}c}
   {94} & 0 & 1  \\
   0 & 5 & 0  \\
\end{array}} \right]$
 \\   \hline
CR & 0.99 & 0.99 & 0.99 & 0.99 & 0.99 & 0.99 & 0.99 & 0.99
\\   \hline
(Rejection) & (0.00) & (0.00) & (0.01) & (0.01) & (0.00) & (0.00) &
(0.01) & (0.01)
\\   \hline
$NI_2$ & 0.756 & 0.874 & 0.876 & 0.997 & 0.720 & 0.864 & 0.849 &
0.997
\\   \hline
(Ranking) & (D) & (C) & (B) & (A) & (D) & (B) & (C) & (A)
\\   \hline
\end{tabular}
\end{center}
\end{table*}
\end{landscape}

\begin{landscape}
\begin{table*}[htb]
\begin{center}
\caption{Classification examples in three
classes($C_1=80$,$C_2=15$,$C_3=5$).(R)= ranking order for the model,
where R = A,B, ..., in descending order from the top.} \small
\begin{tabular}{llllll}
 \hline\noalign{\smallskip}
 Model & M7 & M8 & M9 & M10 & M11 \\
 (Ranking)& (C)& (C)& (B)& (B)& (B) \\
 \hline\noalign{\smallskip}
 \textbf{C} & $
\left[ {\begin{array}{*{8}c}
   {80}&0&0&0  \\
   0&{15}&0&0  \\
   1&0&4&0  \\
\end{array}} \right] $ & $
\left[ {\begin{array}{*{20}c}
   {80} & 0 & 0 & 0  \\
   0 & {15} & 0 & 0  \\
   0 & 1 & 4 & 0  \\
\end{array}} \right] $ & $
\left[ {\begin{array}{*{20}c}
   {80} & 0 & 0 & 0  \\
   0 & {15} & 0 & 0  \\
   0 & 0 & 4 & 1  \\
\end{array}} \right] $ & $
\left[ {\begin{array}{*{20}c}
   {80} & 0 & 0 & 0  \\
   1 & {14} & 0 & 0  \\
   0 & 0 & 5 & 0  \\
\end{array}} \right] $ & $
\left[ {\begin{array}{*{20}c}
   {80} & 0 & 0 & 0  \\
   0 & {14} & 1 & 0  \\
   0 & 0 & 5 & 0  \\
\end{array}} \right]$ \\
\emph{CR} & 0.99 & 0.99 & 0.99& 0.99 & 0.99 \\
\emph{Rej} & 0.00 & 0.00 & 0.01 & 0.00 & 0.00 \\
 \hline
 Model     & M12 & M13 & M14 & M15\\
 (Ranking) & (B) & (B) & (B) & (A)\\
  \hline
 \textbf{C}  & $
\left[ {\begin{array}{*{8}c}
   {80} & 0 & 0 & 0  \\
   0 & {14} & 0 & 1  \\
   0 & 0 & 5 & 0  \\
\end{array}} \right]
$ & $ \left[ {\begin{array}{*{20}c}
   {79} & 1 & 0 & 0  \\
   0 & {15} & 0 & 0  \\
   0 & 0 & 5 & 0  \\
\end{array}} \right] $ & $
\left[ {\begin{array}{*{20}c}
   {79} & 0 & 1 & 0  \\
   0 & {15} & 0 & 0  \\
   0 & 0 & 5 & 0  \\
\end{array}} \right]
$ & $\left[ {\begin{array}{*{20}c}
   {79} & 0 & 0 & 1  \\
   0 & {15} & 0 & 0  \\
   0 & 0 & 5 & 0  \\
\end{array}} \right]$\\
 \emph{CR} & 0.99 & 0.99 & 0.99 & 0.99\\
\emph{Rej} & 0.01 & 0.00 & 0.00 & 0.01\\
\noalign{\smallskip}\hline
\end{tabular}
\end{center}
\end{table*}
\end{landscape}

\begin{landscape}
\begin{table*}[htb]
\begin{center}
\caption{Results for the models in Table 8 on information measures
from mutual-information and cross-entropy groups. S=singularity
which cannot be removed. (R)= ranking order for the model, where R =
A,B, ..., in descending order from the top.} \small
\begin{tabular}{llllllllllllll}
 \hline\noalign{\smallskip}
 Model & $NI_1$ & $NI_2$ & $NI_3$ & $NI_4$ & $NI_5$ & $NI_6$ & $NI_7$ & $NI_8$ & $NI_9$ & $NI_{21}$ & $NI_{22}$& $NI_{23}$& $NI_{24}$\\
 \hline\noalign{\smallskip}
 M7 & 0.912 & 0.912 & 0.957 & 0.935 & 0.934 & 0.934 & 0.876 & 0.912 & 0.957 & 0.998 & 0.998 & 0.998 & 0.998 \\
 (F) & (F) & (F) & (C) & (G) & (G) & (G) & (F) & (H) & (E) & (D) & (C) & (C) & (C)\\
 \hline
 M8 & 0.939 & 0.939 & 0.958 & 0.949 & 0.949 & 0.949 & 0.902 & 0.939 & 0.958 & 0.998 & 0.998 & 0.998 & 0.998\\
 (F) & (E) & (E) & (B) & (D) & (D) & (D) & (D) & (D) & (D) & (D) & (C) & (C) & (C)\\
 \hline
 M9 & 1.000 & 0.951 & 0.961 & 0.980 & 0.980 & 0.980 & 0.961 & 0.961 & 1.000 & 0.982 & 0.000 & 0.491 & 0.000\\
 (C) & (A) & (D) & (A) & (A) & (A) & (A) & (A) & (A) & (A) & (G) & (G) & (I) & (G)\\
 \hline
 M10 & 0.912 & 0.912 & 0.938 & 0.925 & 0.925 & 0.925 & 0.860 & 0.912 & 0.938 & 0.999 & 0.999 & 0.999 & 0.999\\
 (E) & (F) & (F) & (F) & (I) & (I) & (I) & (H) & (H) & (G) & (A) & (A) & (A) & (A)\\
 \hline
 M11 & 0.956 & 0.956 & 0.941 & 0.948 & 0.948 & 0.948 & 0.902 & 0.941 & 0.956 & 0.998 & 0.998 & 0.998 & 0.998\\
 (E) & (D) & (C) & (E) & (E) & (E) & (E) & (D) & (C) & (E) & (B) & (C) & (C) & (C)\\
 \hline
 M12 & 1.000 & 0.969 & 0.943 & 0.972 & 0.971 & 0.971 & 0.943 & 0.943 & 1.000 & 0.983 & 0.000 & 0.492 & 0.000\\
 (B) & (A) & (B) & (D) & (B) & (B) & (B) & (B) & (B) & (A) & (F) & (G) & (G) & (G)\\
 \hline
 M13 & 0.939 & 0.939 & 0.915 & 0.927 & 0.927 & 0.927 & 0.863 & 0.915 & 0.939 & 0.999 & 0.999 & 0.999 & 0.999\\
 (D) & (E) & (E) & (I) & (H) & (H) & (H) & (G) & (G) & (F) & (A) & (A) & (A) & (A)\\
 \hline
 M14 & 0.956 & 0.956 & 0.916 & 0.936 & 0.935 & 0.936 & 0.879 & 0.916 & 0.956 & 0.998 & 0.998 & 0.998 & 0.998\\
 (D) & (D) & (C) & (H) & (F) & (F) & (F) & (E) & (F) & (E) & (D) & (C) & (C) & (C)\\
 \hline
 M15 & 1.000 & 0.996 & 0.919 & 0.960 & 0.958 & 0.959 & 0.919 & 0.919 & 1.000 & 0.984 & 0.000 & 0.492 & 0.000\\
 (A) & (A) & (A) & (G) & (C) & (C) & (C) & (C) & (E) & (A) & (E) & (G) & (G) & (G)\\
 \noalign{\smallskip}\hline
\end{tabular}
\end{center}
\end{table*}
\end{landscape}

\begin{landscape}
\begin{table*}[htb]
\begin{center}
\caption{Results for the models in Table 8 on information measures
from divergence group. S=singularity which cannot be removed. (R)=
ranking order for the model, where R = A,B, ..., in descending order
from the top.} \small
\begin{tabular}{llllllllllll}
 \hline\noalign{\smallskip}
  Model & $NI_{10}$ & $NI_{11}$ & $NI_{12}$ & $NI_{13}$ & $NI_{14}$ & $NI_{15}$ & $NI_{16}$ & $NI_{17}$ & $NI_{18}$ & $NI_{19}$ & $NI_{20}$\\
\hline\noalign{\smallskip}
  M7 & 0.9998 & 0.9998 & 0.9982 & 0.9996 & 0.9974 & 0.9994 & 0.9802 & 0.9966 & 0.9992 & 0.9953 & 0.9992\\
  (F) & (A) & (A) & (D) & (C) & (E) & (D) & (A) & (D) & (D) & (E) & (D)\\
\hline
  M8 & 0.9998 & 0.9996 & 0.9979 & 0.9995 & 0.9969 & 0.9993 & 0.9802 & 0.9959 & 0.9990 & 0.9942 & 0.9990\\
  (F) & (A) & (E) & (E) & (D) & (F) & (E) & (A) & (F) & (F) & (F) & (F)\\
\hline
  M9 & 0.9998 & 0.9996 & 0.9840 & 0.9924 & 0.9876 & 0.9895 & 0.9802 & S & 0.9893 & S & S\\
  (C) & (A) & (E) & (H) & (G) & (I) & (H) & (A) &  & (H) &  & \\
\hline
  M10 & 0.9998 & 0.9997 & 0.9994 & 0.9999 & 0.9992 & 0.9998 & 0.9802 & 0.9988 & 0.9997 & 0.9984 & 0.9997\\
  (E) & (A) & (C) & (A) & (A) & (A) & (A) & (A) & (B) & (A) & (C) & (A)\\
 \hline
  M11 & 0.9998 & 0.9996 & 0.9982 & 0.9995 & 0.9976 & 0.9994 & 0.9802 & 0.9964 & 0.9991 & 0.9950 & 0.9991\\
  (E) & (A) & (E) & (D) & (D) & (D) & (D) & (A) & (E) & (E) & (F) & (E)\\
  \hline
  M12 & 0.9998 & 0.9996 & 0.9852 & 0.9927 & 0.9893 & 0.9899 & 0.9802 & S & 0.9898 & S & S \\
  (B) & (A) & (E) & (G) & (F) & (H) & (G) & (A) &  & (H) &  & \\
  \hline
  M13 & 0.9998 & 0.9997 & 0.9994 & 0.9999 & 0.9992 & 0.9998 & 0.9802 & 0.9989 & 0.9997 & 0.9985 & 0.9997\\
  (D) & (A) & (C) & (A) & (A) & (A) & (A) & (A) & (A) & (A) & (A) & (A)\\
  \hline
  M14 & 0.9998 & 0.9997 & 0.9986 & 0.9996 & 0.9982 & 0.9995 & 0.9802 & 0.9972 & 0.9993 & 0.9961 & 0.9993\\
  (D) & (A) & (C) & (C) & (C) & (C) & (C) & (A) & (C) & (C) & (D) & (C)\\
  \hline
  M15 & 0.9998 & 0.9998 & 0.9856 & 0.9928 & 0.9899 & 0.9900 & 0.9802 & S & 0.9900 & S & S\\
  (A) & (A) & (A) & (F) & (E) & (G) & (F) & (A) &  & (G) &  & \\
  \noalign{\smallskip}\hline
\end{tabular}
\end{center}
\end{table*}
\end{landscape}

\end{document}